\documentclass[10pt,twocolumn,letterpaper]{article}

\usepackage[pagenumbers]{cvpr}

\ifdefined\pdfsuppressptexinfo \pdfsuppressptexinfo=-1 \fi

\usepackage{amsmath}
\usepackage{amssymb}
\usepackage{booktabs}
\usepackage{graphicx}
\usepackage{multirow}
\usepackage{microtype}

\makeatletter
\newsavebox{\cvpr@teaserbox}
\newif\ifcvpr@teaser
\newenvironment{teaserfigure}
  {\begin{lrbox}{\cvpr@teaserbox}\begin{minipage}{\textwidth}%
   \captionsetup{type=figure,hypcap=false}}
  {\end{minipage}\end{lrbox}%
   \global\setbox\cvpr@teaserbox=\box\cvpr@teaserbox
   \global\cvpr@teasertrue}
\apptocmd{\@maketitle}{%
  \ifcvpr@teaser
    \begin{center}\usebox{\cvpr@teaserbox}\end{center}
  \fi
}{}{\PackageError{preamble}{Cannot patch \string\@maketitle}{}}
\makeatother

\renewcommand{\abstract}{%
   \iftoggle{cvprpagenumbers}{}{%
     \thispagestyle{empty}
   }
   \centerline{\large\bf Abstract}%
   \vspace*{6pt}\noindent%
   \it\ignorespaces%
}

\definecolor{cvprblue}{rgb}{0.21,0.49,0.74}
\usepackage[pagebackref,breaklinks,colorlinks,allcolors=cvprblue]{hyperref}
\usepackage[capitalize,nameinlink]{cleveref}

\def\confName{CVPR}
\def\confYear{2027}

\title{UniCross: Unified Cross-Skill Dexterous Manipulation Synthesis}
\author{
Hui Zhang\textsuperscript{1,2}
\thanks{Correspondence: huizhang@ethz.ch}
\thanks{Project page: \url{https://zdchan.github.io/UniCross/}}
\quad
Julian Ferchow\textsuperscript{2} \quad
Jie Song\textsuperscript{3} \quad
Mirko Meboldt\textsuperscript{1}\\
\vspace{-2mm}
\\
{\normalsize
\textsuperscript{1}ETH Z\"urich, Switzerland \quad
\textsuperscript{2}inspire AG, Z\"urich, Switzerland \quad
\textsuperscript{3}HKUST (Guangzhou), China
}
}

\begin{document}
\begin{teaserfigure}
    \centering
    \vspace{-6mm}%
    \includegraphics[width=1.0\textwidth]{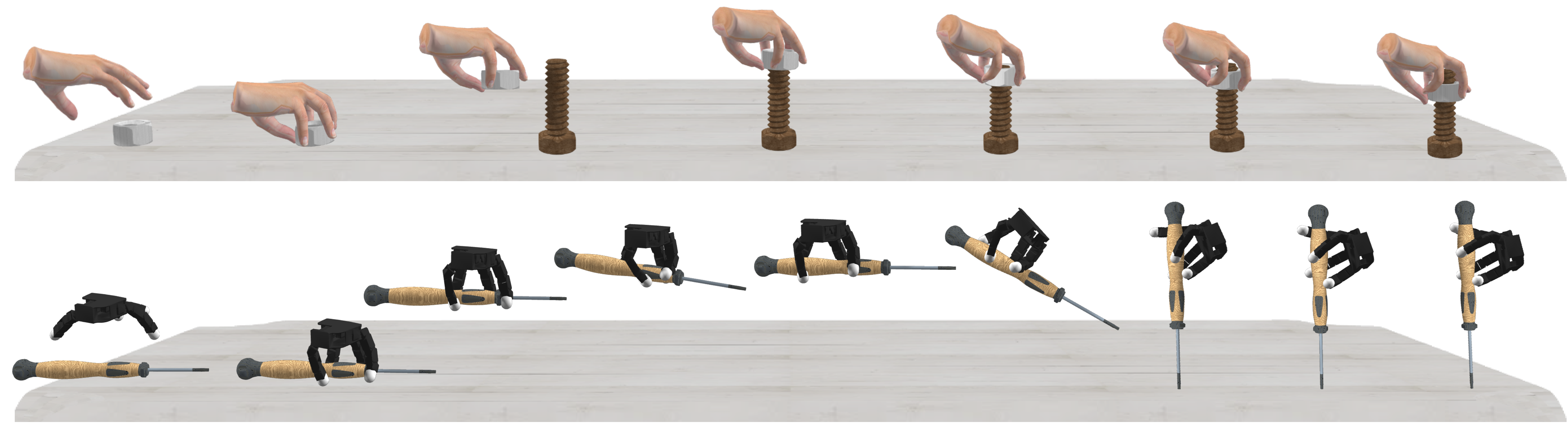}
    \vspace{-7mm}
    \caption{We present a unified dexterous manipulation framework that models grasping, relocation, in-hand translation, and in-hand rotation under a shared hand-object relational formulation. This coherent formulation facilitates cross-skill compatibility and continuity, enabling a single policy to perform long-horizon manipulation through seamless skill chaining based on relational targets. The embodiment-agnostic framework transfers effectively across different hand morphologies.}
    \vspace{1.5mm}
    \label{fig:teaser}
    \end{teaserfigure}

\maketitle

\begin{abstract}
  Many dexterous manipulation tasks require the object to remain securely held throughout the interaction. From the perspective of hand-object relational motion, such manipulation comprises four canonical skills: grasping, relocation, in-hand rotation, and in-hand translation. Human hands flexibly compose these skills to accomplish complex tasks. Existing approaches, however, model these skills separately with skill-specific action constraints, objectives, or even dedicated hand morphologies, which breaks the compatibility and continuity required for long-horizon composition. In this work, we present a unified framework that models all four skills in a single formulation instantiated with desired relational targets. The formulation preserves cross-skill compatibility and enables distillation of a single policy without skill identifiers, which achieves strong per-skill performance and generalizes to unseen objects. The compatibility further supports the policy to perform long-horizon skill chaining and execute relational targets unseen during training. The framework also applies across different hand morphologies. Together, these results suggest that distinct dexterous manipulation skills can be modeled as instantiations of a shared relational task formulation.
\end{abstract}

\section{Introduction}
\label{sec:introduction}

Human hands exhibit remarkable dexterity and agility. Through diverse contact modulation, they can establish stable grasps, reposition objects, and perform in-hand adjustments while keeping the object securely held by the hand.
With the capability of maintaining object stability while preserving operational flexibility, this kind of manipulation is the foundation for complex downstream tasks such as plug insertion, screw tightening, and cutting with a knife, which are common in natural human activities. As a result, modeling such manipulation motions is meaningful in different domains, including animation, AR/VR, and robotics.

From the perspective of hand-object relational motion, this kind of manipulation comprises four canonical behaviors: grasping, relocation, in-hand rotation, and in-hand translation. Together, these behaviors form fundamental building blocks of everyday human actions and collectively enable human dexterity.
Prior works in dexterous manipulation have primarily focused on individual skills with specific settings, such as encouraging persistent large contact forces around static grasping poses \cite{christen2022dgrasp}, performing in-hand translation with fixed palm-up wrist poses \cite{yin2024learninginhandtranslation}, or utilizing a specific hand morphology for in-hand rotation \cite{chen2023visual}.
Human manipulation, in contrast, usually involves the continuous composition and seamless transition of these skills, rather than their isolated execution. This motivates a unified formulation that captures the shared structures and intrinsic consistency across different dexterous manipulation skills, instead of relying on skill-specific designs that disrupt the compatibility and continuity for skill composition.

A cross-skill unified formulation for dexterous manipulation, however, is challenging in several aspects.
First, these skills require fundamentally different contact regimes. Grasping and relocation favor persistent, stable contacts, while in-hand rotation and translation often demand frequent contact reconfiguration to balance stability and mobility. This diversity makes it challenging to unify state spaces, reward structures, and other task definitions across skills.
Second, unifying different skills imposes higher requirements on object shape representation. Grasping needs to identify shape regions that can support stable contact forces, while in-hand manipulation requires dynamically capturing shape features to adjust finger poses according to motion objectives. A unified representation must capture these diverse shape characteristics across all skills, rather than resorting to skill-specific heuristics or object-dependent tuning.
Third, seamless skill composition for long-horizon manipulation requires compatibility across skills. States reached by one skill should remain valid for another, which makes skill-specific settings (palm-up in-hand manipulation with palm support, skill-specific hand morphologies, etc.) infeasible.

In this work, we take a step toward unified dexterous manipulation synthesis. Instead of defining skills based on contact regimes, we adopt a different perspective centered on hand-object relational motion. This perspective is inherently unifying, as all four skills fundamentally aim to realize desired hand-object relational motions while maintaining object stability, regardless of their specific contact patterns.
Guided by this view, we model different skills as different instantiations of a single formulation conditioned on desired relational motions, while sharing state representations, action spaces, and reward structures.
To capture object shapes, we adopt the hand-centric representation from \cite{zhang2024graspxl}, originally proposed for grasping, and show that it remains effective across all four skills by dynamically tracking interaction-relevant contact states and local geometric features.
With each skill instantiated by continuous relational targets rather than a discrete identifier, our framework preserves cross-skill compatibility, allowing straightforward distillation of a single policy that not only chains skills into long-horizon manipulation, but also executes relational targets unseen in training, \eg, rotating objects about arbitrary axes.

Overall, our contributions can be summarized as follows:
\begin{itemize}
    \item We introduce a hand-object relational perspective for dexterous manipulation synthesis, under which grasping, relocation, in-hand rotation, and in-hand translation are modeled with a single formulation, instantiated by the desired relational targets.
    \item The formulation preserves cross-skill compatibility, allowing straightforward distillation of a single policy to execute all four skills and chain them into long-horizon manipulation without explicit skill identifiers.
    \item We demonstrate that the policy trained with our unified formulation achieves strong per-skill performance
    and generalization to unseen objects. The framework also transfers across hand morphologies.
    \item More broadly, the unified policy executes relational targets unseen in training, suggesting that different dexterous manipulation skills can be modeled as instantiations of a shared solution space rather than isolated behaviors.
\end{itemize}

\section{Related Work}

\subsection{Hand-object Interaction Synthesis}

Hand-object interaction (HOI) synthesis has long been studied in computer graphics and vision~\cite{rijpkema1991computer, chen2023synthesizing, guitar, pang2025manivideo, wu2022saga, braun2024physically, jiang2021graspTTA, kalisiak2001grasp, karunratanakul2020grasping, wang2025learning}.
Prior work usually formulated HOI synthesis as optimization problems. Specifically, some works focus on utilizing existing static grasping poses to generate new poses or animated motions, such as generating grasping poses for new objects by finetuning similar poses retrieved from a database~\cite{Li2007DataDriven, chen2025dexonomy}, generating finger motions by fitting given reference poses~\cite{Pollard2005PhysicallyBased}, or fitting a partial object motion with an initial grasp~\cite{Liu2009Dextrous}. To better handle physical plausibility, some work explicitly incorporated contact into the synthesis pipeline by fitting captured contact forces for other objects~\cite{kry2006interaction}, sampling contact-consistent trajectories~\cite{Ye2012DetailedHand}, performing contact-invariant optimization~\cite{Mordatch2012Contact, Multifinger2013}, or optimizing compliant contact forces under shape uncertainty~\cite{chen2024springgrasp}.
While effective for specific manipulation problems, these optimization-based methods are usually computationally expensive, and rely on per-task objectives such as reference poses, contact modes, or annotated grasp templates~\cite{brahmbhatt2019contactdb}. Furthermore, these methods usually simplify physical constraints (e.g., no object gravity~\cite{Multifinger2013}), leading to unrealistic motion behaviors.

Instead of relying on specific objectives, some recent work utilizes data-driven learning approaches that train generative models \cite{zhang2025bimart, Zheng_2023_CAMS, taheri2024grip, liu2024geneoh, zhao2013robust, cha2024text2hoi, peng2025hoi} by extracting priors from captured hand-object motion datasets~\cite{taheri2020grab, fan2023arctic, chao2021dexycb, liu2024taco, zhan2024oakink2}.
While these methods can synthesize realistic motions within the data distribution, their generalization is fundamentally constrained by dataset coverage, particularly when transferring to unseen object geometries or hand embodiments with different morphologies. Some of these works require reference trajectories during inference (e.g., wrist or object motions~\cite{li2025maniptrans, zhang2021manipnet}), which limits their ability to synthesize novel manipulation behaviors. Moreover, without explicit physics modeling, these methods usually suffer from physical implausibility such as interpenetration or jittering~\cite{taheri2021goal, ghosh2023imos}.

Another line of work learns dexterous manipulation in simulation via reinforcement learning (RL)~\cite{rajeswaran2018learning, christen2022dgrasp, luo2024omnigrasp, zhang2024artigrasp, she2022learning, yang2022learning, zhang2025RobustDexGrasp, huang2024fungrasp}. By leveraging physics-driven exploration, these methods offer stronger physical grounding while alleviating the constraints imposed by limited real-world data.
However, due to the difficulty of extracting shared representations and unified objectives across different manipulation behaviors, existing RL methods typically focus on individual skills and rely heavily on task-specific heuristics, reward shaping, architectures, or exploration constraints, such as restricted action spaces~\cite{lum2024dextrahg}, specific hand-object configurations~\cite{qi2023general, yin2024learninginhandtranslation}, or specialized hand morphologies~\cite{chen2023visual}.
These skill-specific constraints break the continuity of hand-object states across skills and make skill composition infeasible.
Our work, following the RL-in-simulation paradigm, departs from skill-specific designs by formulating different dexterous manipulation behaviors from a unified cross-skill hand-object relational perspective.

\subsection{Multi-Skill and Generalist Policy Learning}
Generalist policy learning has shown promising results in character animation. Based on a unified solution space which models different skills, a single controller can synthesize diverse whole-body behaviors, such as different locomotion modes~\cite{peng2022ase, MCPPeng19, tessler2023calm, juravsky2024superpadl} and complex scene interactions~\cite{pan2025tokenhsi, tessler2024maskedmimic, xu2025intermimic}.
Dexterous manipulation, however, is still at an early stage of generalist policy learning.
The fundamental challenge is that seemingly small changes in object geometry, contact configuration, or hand morphology can induce largely different manipulation strategies, making it difficult to extract shared motion primitives. As a result, it is still unclear whether a unified solution space exists for dexterous manipulation, and whether it can be modeled with a shared formulation.

Due to this challenge, prior work on dexterous manipulation mainly focuses on individual skills with specific contact patterns.
Most research still centers on grasping \cite{christen2024diffh2o, zhou2024gears, zhang2024graspxl, Muchen_LatentHOI, she2022learning, wan2023unidexgraspplus}, as the consistent contact makes it easier to model and formulate.
Recently, some works extend dexterous manipulation from grasping to other skills such as articulation \cite{zhang2025bimart, zhang2024artigrasp, SUN2026112783, zhang2025manidext, Zheng_2023_CAMS}. However, these works depend heavily on existing articulated datasets \cite{fan2023arctic, liu2022hoi4d} and do not generalize to unseen shapes. Moreover, articulation primarily involves wrist-level motion under a fixed grasp, while fine-grained in-hand manipulation with coordinated finger motion remains unexplored.

Several works in the robotics domain explore in-hand rotation \cite{chen2023visual, qi2023general, liu2025dexndm} and translation \cite{yin2024learninginhandtranslation}, but they are still restricted to single-skill policies and often rely on highly specific assumptions. For example, Yin et al. \cite{yin2024learninginhandtranslation} rely on palm support for object in-hand translation, which requires hand-up configurations, while Chen et al. \cite{chen2023visual} utilize a symmetric-finger hand morphology to rotate objects with fixed wrist poses. Such assumptions limit cross-skill compatibility and prevent skill composition.

A few recent works have different focuses. Instead of generating specific motions, Dex4D~\cite{kuang2026dex4d} and SimToolReal~\cite{kedia2026simtoolreal} bring an object to world-frame target poses or point tracks, which specify where the object goes but not how,
so the behavior is emergent and a specific skill cannot be requested.
DexterityGen~\cite{yin2025dexteritygen} is a low-level finger controller driven by external fingertip commands, \eg, from teleoperation, with the wrist commanded separately. Our work, in contrast, focuses on unifying the synthesis of different types of hand-object relational motion behaviors.

\section{Method}
\label{sec:method}

\begin{figure*}
    \centering
    \includegraphics[width=1.0\textwidth]{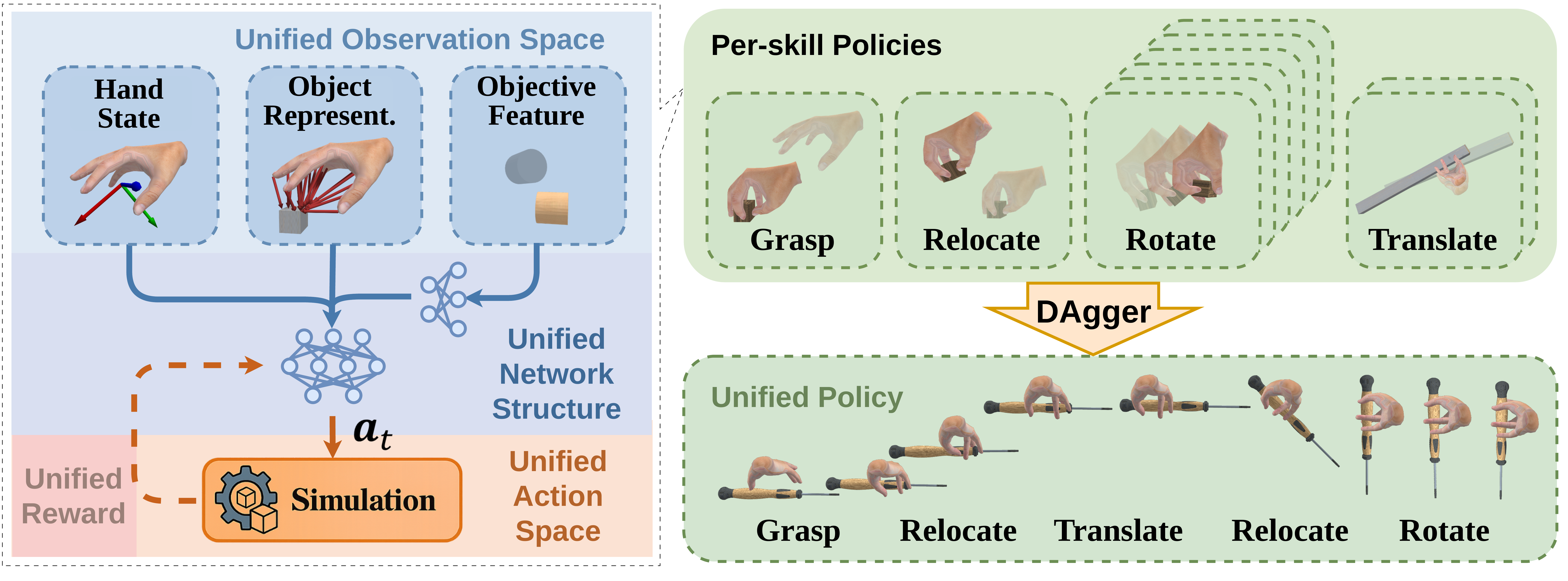}
    \vspace{-6mm}
    \caption{Overview of the unified framework. All skills share a relational representation, network architecture, action space, and reward formulation, and are instantiated by their relational targets (\cref{tab:target_poses}). Ten per-skill policies (one for grasp, one for relocate, six for rotation (x+/x-/y+/y-/z+/z- of the hand frame), and two for translation (z+/z- of the hand frame)) are trained under this formulation. The experts are then distilled into a single policy conditioned on relational targets without skill identifiers.
    }
    \label{fig:pipeline}
\end{figure*}

This work aims to unify the formulation of different dexterous manipulation skills within a single framework, supporting cross-skill coherence. To this end, we formulate the four skills from a shared hand-object relational perspective.
As illustrated in \cref{fig:pipeline}, we first train per-skill policies with PPO \cite{schulman2017proximal}, which share a unified observation space, network architecture, action space, and reward structure (\cref{sec:task_formulation}). The unified representation is instantiated for each policy based on the relational-motion-relevant objectives (\cref{sec:implementation}). With cross-skill compatibility, the per-skill policies can be straightforwardly distilled into a single policy with vanilla DAgger \cite{ross2011dagger} (\cref{sec:distillation}). All policies are trained in IsaacGym \cite{makoviychuk2021isaacgym}, with a simulation frequency of 120 Hz and a control frequency of 20 Hz.
More details are in SuppMat.

\subsection{Task Formulation}
\label{sec:task_formulation}

\subsubsection{Task Definition}
\label{sec:task_definition}
Given a dexterous hand model and a rigid object, the framework trains policies in physics simulation to control the hand and generate diverse hand-object interaction motions.
To model the hand, we first define a root frame at the beginning of each episode using the initial wrist pose in the world frame, and keep it fixed thereafter for the entire episode. We then parameterize the wrist motion with six zero-initialized virtual joints, including three translational and three rotational joints. This formulation represents wrist motion relative to the initial pose instead of absolute world coordinates, which benefits generalization.
On top of this, we define the hand frame using the current hand pose (see Hand State in \cref{fig:pipeline}). The object pose is then represented in the hand frame by $\mathbf{x}_{o}^{h}$ and $\mathbf{q}_{o}^{h}$, and in the root frame by $\mathbf{x}_{o}^{r}$ and $\mathbf{q}_{o}^{r}$. The hand pose in the root frame is represented by $\mathbf{x}_{h}^{r}$ and $\mathbf{q}_{h}^{r}$.
Target poses are indicated by $\hat{\cdot}$, including $\hat{\mathbf{x}}_{o}^{h}$, $\hat{\mathbf{q}}_{o}^{h}$, $\hat{\mathbf{x}}_{o}^{r}$, $\hat{\mathbf{q}}_{o}^{r}$, $\hat{\mathbf{x}}_{h}^{r}$, and $\hat{\mathbf{q}}_{h}^{r}$.
With these definitions, each manipulation skill can be characterized from the perspective of hand-object relational movements in two frames:
\begin{itemize}
    \item \textbf{Grasp:} the object is fixed in the root frame, while the hand moves in the root frame towards the object.
    \item \textbf{Relocate:} the object is fixed in the hand frame, while the wrist moves in the root frame to reach a target object pose.
    \item \textbf{Rotate:} the wrist is fixed in the root frame, the object position is fixed in the hand frame, and the object orientation rotates about a designated axis in the hand frame.
    \item \textbf{Translate:} the wrist is fixed in the root frame, the object orientation is fixed in the hand frame, and the object position translates along a designated axis in the hand frame.
\end{itemize}

\subsubsection{Observation Space}
\label{sec:observation_space}
The observation space $\textbf{o}_t$ provides the representation of the current hand-object states, interaction-relevant object shape features shared across skills, and the hand-object relational motion objectives. It is designed as $\textbf{o}_t = (\mathbf{s}_t^{\text{h}}, \mathbf{s}_t^{\text{o}}, \mathbf{g}_t)$, where $\mathbf{s}_t^{\text{h}}$ is the hand state, $\mathbf{s}_t^{\text{o}}$ is the interaction-aware object representation, and $\mathbf{g}_t$ is the relation-based objective features.

Specifically, $\mathbf{s}_t^{\text{h}} = [\mathbf{q}_t, \mathbf{q}_{t-1}^{\text{target}}]$ includes the current positions and previous target positions of all hand joints (finger joints and virtual wrist joints).
The object is represented with $\mathbf{s}_t^{\text{o}} = [\mathbf{c}_t, \mathbf{f}_t, \mathbf{v}_t]$, including the per-link contact states $\mathbf{c}_t$, contact force magnitudes $\mathbf{f}_t$, and the distance vectors $\mathbf{v}_t$ from each finger link to their nearest points on the object surface. This interaction-centric representation, inspired by \cite{zhang2024graspxl}, captures hand-object spatial relationships as well as object geometry features shared across manipulation skills.
The policies receive a 39-dimensional relational feature vector $\mathbf{g}_t = [\mathbf{g}_t^{\text{current}},\mathbf{g}_t^{\text{target}}]$, with the current multi-frame relational poses $\mathbf{g}_t^{\text{current}} = [\mathbf{q}_{h}^{r}, \mathbf{x}_{o}^{h}, \mathbf{q}_{o}^{h}, \mathbf{x}_{o}^{r}, \mathbf{q}_{o}^{r}]$ and their target poses $\mathbf{g}_t^{\text{target}} = [\hat{\mathbf{x}}_{o}^{h}, \hat{\mathbf{q}}_{o}^{h}, \hat{\mathbf{x}}_{o}^{r}, \hat{\mathbf{q}}_{o}^{r}, \hat{\mathbf{x}}_{h}^{r}, \hat{\mathbf{q}}_{h}^{r}]$ that poses are instantiated in \cref{sec:objective-relevant-pose} according to the motion objectives in \cref{sec:task_definition}.

\subsubsection{Action Space}
\label{sec:action_space_and_control}

The action space parameterizes the full hand motion, including six virtual wrist joints and all articulated finger joints.
The policy outputs incremental motion commands $\mathbf{a}_t$ that are transformed to target joint positions with
$\mathbf{q}_t^{\text{act}} = \text{clamp}(\mathbf{q}_{t}^{\text{ref}} + \boldsymbol{\alpha} \cdot \mathbf{a}_t, \mathbf{q}_{\min}, \mathbf{q}_{\max})$,
where $\boldsymbol{\alpha}$ is the action scale and $\text{clamp}(\cdot)$ enforces anatomical joint limits defined by $\mathbf{q}_{\min}$ and $\mathbf{q}_{\max}$.
Specifically, $\mathbf{q}_{t}^{\text{ref}}$ is set to $\mathbf{q}_{t-1}^{\text{act}}$ for the finger joints and to the current target wrist pose ($\hat{\mathbf{x}}_{h}^{r}$ and the Euler angles of $\hat{\mathbf{q}}_{h}^{r}$) for the virtual wrist joints. This incremental motion parameterization facilitates temporally coherent and smooth finger motions while accelerating wrist motion exploration.
Target joint positions $\mathbf{q}_t^{\text{act}}$ are then converted to joint torques by a PD controller, which are applied to the joints to generate the desired motion in simulation.

\subsubsection{Reward Design}
From the shared hand-object relational perspective, we define a set of reward terms to encourage efficient and effective completion of different manipulation skills:

\begin{equation}
r_t = r_t^{\text{goal}} + r_t^{\text{track}} + r_t^{\text{reg}}
\end{equation}

where $r_t^{\text{goal}}$ encodes general hand-object interaction objectives, $r_t^{\text{track}}$ encourages tracking of the hand-object target poses $\mathbf{g}_t^{\text{target}}$ defined in \cref{sec:observation_space}, and $r_t^{\text{reg}}$ promotes stable and efficient manipulation via a set of penalty terms.

The object should remain securely held by the hand while performing desired movements. Motivated by this, the goal term $r_t^{\text{goal}}$ is designed to encourage stable hand-object contact and object motion along the target axis:
\begin{equation}
r_t^{\text{goal}} = r_t^{\text{contact}} + r_t^{\text{motion}}
\end{equation}
The contact term is defined as $r_t^{\text{contact}} = \frac{1}{L} \sum_{i=1}^{L}(-w_{\text{dis}} \cdot \delta_i + w_{\text{con}} \cdot c_i + w_{\text{f}} \cdot f_i)$, where $L$ is the number of hand links, $\delta_i$ is the distance from link $i$ to the object, $c_i \in \{0, 1\}$ indicates whether link $i$ is in contact, and $f_i$ is the contact force magnitude.
The motion term is $r_t^{\text{motion}} = w_{\text{p}} \cdot \min(\boldsymbol{\nu}_o^{h} \cdot \mathbf{d}^{\text{h}}, \nu_{\max})$, where $\mathbf{d}^{\text{h}}$ is the target moving direction in the hand frame, and $\boldsymbol{\nu}_o^{h}$ is the object hand-frame linear or angular velocity corresponding to the motion type, and the threshold $\nu_{\max}$ limits excessive motion.

The target tracking component $r_t^{\text{track}}$ penalizes hand and object deviations from target poses in root and hand frames, with $\mathcal{A}(\cdot, \cdot)$ denoting the quaternion angular difference:
\begin{equation}
r_t^{\text{track}} =
\begin{aligned}[t]
& w_{opr} \|\mathbf{x}_{o,t}^{r} - \hat{\mathbf{x}}_{o}^{r}\|^2
+ w_{oqr} \mathcal{A}(\mathbf{q}_{o,t}^{r}, \hat{\mathbf{q}}_{o}^{r}) \\
& + w_{oph} \|\mathbf{x}_{o,t}^{h} - \hat{\mathbf{x}}_{o}^{h}\|^2
+ w_{oqh} \mathcal{A}(\mathbf{q}_{o,t}^{h}, \hat{\mathbf{q}}_{o}^{h}) \\
& + w_{hpr} \|\mathbf{x}_{h,t}^{r} - \hat{\mathbf{x}}_{h}^{r}\|^2
+ w_{hqr} \mathcal{A}(\mathbf{q}_{h,t}^{r}, \hat{\mathbf{q}}_{h}^{r}) .
\end{aligned}
\end{equation}

The regularization component $r_t^{\text{reg}}$ encourages stable and efficient manipulation through multiple penalty terms:
\begin{equation}
r_t^{\text{reg}} = r_t^{\text{pose}} + r_t^{\text{vel}} + r_t^{\text{energy}} + r_t^{\text{drop}}
\end{equation}
where $r_t^{\text{pose}} = w_{\text{pose}} \sum_{i=1}^{J} (q_{i,t} - q_{i,0})^2$ penalizes deviations from the initial finger joint positions to prevent excessive finger movements, with the finger joint number $J$ and the initial joint position $q_{i,0}$.
The wrist velocity penalty $r_t^{\text{vel}} = w_{\text{vel}} \|\dot{\mathbf{x}}_{h,t}\|^2 + w_{\text{ang}} \|\boldsymbol{\omega}_{h,t}\|^2$ penalizes wrist translational and angular velocities $\dot{\mathbf{x}}_{h,t}$ and $\boldsymbol{\omega}_{h,t}$, promoting smooth wrist motion. The energy penalty
$r_t^{\text{energy}} = w_{\tau} \|\boldsymbol{\tau}_t\|^2$
penalizes the magnitude of applied joint torques $\boldsymbol{\tau}_t$, encouraging energy-efficient control.
The drop penalty $r_t^{\text{drop}} = w_{\text{drop}} \mathbb{I}_{\text{drop}}$ penalizes the early termination caused by object dropping, with the indicator $\mathbb{I}_{\text{drop}}$ defined by whether the distance between all hand links and the object exceeds a threshold.

All the reward weights and other hyperparameters are provided in SuppMat.

\begin{table}[t]
\centering
\caption{Instantiation of the target poses $\mathbf{g}_t^{\text{target}}$ for each skill.}
\vspace{-1mm}
\label{tab:target_poses}
\resizebox{0.9\columnwidth}{!}{
\begin{tabular}{l|c|c|c}
    \toprule
    Skill & Tracked & Fixed & Free \\
    \midrule
    Grasp & $\hat{\mathbf{x}}_{o}^{h}$ & $\hat{\mathbf{x}}_{o}^{r}, \hat{\mathbf{q}}_{o}^{r}$ & $\hat{\mathbf{q}}_{o}^{h}, \hat{\mathbf{x}}_{h}^{r}, \hat{\mathbf{q}}_{h}^{r}$ \\
    Relocate & $\hat{\mathbf{x}}_{o}^{r}, \hat{\mathbf{q}}_{o}^{r}, \hat{\mathbf{x}}_{h}^{r}, \hat{\mathbf{q}}_{h}^{r}$ & $\hat{\mathbf{x}}_{o}^{h}, \hat{\mathbf{q}}_{o}^{h}$ & None \\
    Rotate & $\hat{\mathbf{q}}_{o}^{h}, \hat{\mathbf{q}}_{o}^{r}$ & $\hat{\mathbf{x}}_{h}^{r}, \hat{\mathbf{q}}_{h}^{r}, \hat{\mathbf{x}}_{o}^{r}, \hat{\mathbf{x}}_{o}^{h}$ & None \\
    Translate & $\hat{\mathbf{x}}_{o}^{h}, \hat{\mathbf{x}}_{o}^{r}$ & $\hat{\mathbf{x}}_{h}^{r}, \hat{\mathbf{q}}_{h}^{r}, \hat{\mathbf{q}}_{o}^{r}, \hat{\mathbf{q}}_{o}^{h}$ & None \\
    \bottomrule
\end{tabular}
}
\vspace{-2mm}
\end{table}

\subsection{Relational Target Pose Instantiation}
\label{sec:implementation}
\label{sec:objective-relevant-pose}

The relational target, $\mathbf{g}_t^{\text{target}}$, contains the target hand and object poses in the
root and hand frames: $\hat{\mathbf{x}}_{o}^{r}$, $\hat{\mathbf{q}}_{o}^{r}$, $\hat
{\mathbf{x}}_{o}^{h}$, $\hat{\mathbf{q}}_{o}^{h}$, $\hat{\mathbf{x}}_{h}^{r}$, and
$\hat{\mathbf{q}}_{h}^{r}$, which instantiates the motion objectives defined in
\cref{sec:task_definition}.
As summarized in \cref{tab:target_poses}, each target is set as \textit{Tracked} (dynamically updated toward the skill objective), \textit{Fixed} (held at the initial variable value), or \textit{Free} (set to the current value so the corresponding DOF is unconstrained).

For grasping, the object should maintain its initial root-frame pose ($\hat{\mathbf{x}}_{o}^{r} = \mathbf{x}_{o,0}^{r}$, $\hat{\mathbf{q}}_{o}^{r} = \mathbf{q}_{o,0}^{r}$), with the target position in the hand frame set to the origin ($\hat{\mathbf{x}}_{o}^{h} = \mathbf{x}_{\text{hand}}$). The other variables are set to current values for free movements: $\hat{\mathbf{q}}_{o}^{h} = \mathbf{q}_{o}^{h}$, $\hat{\mathbf{x}}_{h}^{r} = \mathbf{x}_{h}^{r}$, $\hat{\mathbf{q}}_{h}^{r} = \mathbf{q}_{h}^{r}$.

For relocation, the object should be stationary in the hand frame ($\hat{\mathbf{x}}_{o}^{h} = \mathbf{x}_{o,0}^{h}$, $\hat{\mathbf{q}}_{o}^{h} = \mathbf{q}_{o,0}^{h}$). With a given object final 6D pose $\mathbf{T}^\text{final}$, the target root-frame object pose $\hat{\mathbf{x}}_{o}^{r}$ and $\hat{\mathbf{q}}_{o}^{r}$ are calculated by linearly interpolating between the current pose $\mathbf{x}_{o}^{r}$ and $\mathbf{q}_{o}^{r}$ towards $\mathbf{T}^\text{final}$, and the root-frame wrist targets $\hat{\mathbf{x}}_{h}^{r}$ and $\hat{\mathbf{q}}_{h}^{r}$ are calculated from $\hat{\mathbf{x}}_{o}^{r}$, $\hat{\mathbf{q}}_{o}^{r}$ by maintaining the current object hand-frame pose $\mathbf{x}_{o}^{h}$ and $\mathbf{q}_{o}^{h}$.

For rotation, the wrist should be fixed in the root frame ($\hat{\mathbf{x}}_{h}^{r} = \mathbf{x}_{h,0}^{r}$, $\hat{\mathbf{q}}_{h}^{r} = \mathbf{q}_{h,0}^{r}$), and the target object orientation in the hand frame is dynamically updated by rotating the current orientation around the target direction $\mathbf{d}^{\text{h}}$: $\hat{\mathbf{q}}_{o,t}^{h} = \mathbf{R}(\omega_{\max} \Delta t, \mathbf{d}^{\text{h}}) \cdot \mathbf{q}_{o,t}^{h}$, with $\hat{\mathbf{q}}_{o}^{r}$ derived from the current wrist orientation $\mathbf{q}_{h}^{r}$. The object position maintains initial values in both frames ($\hat{\mathbf{x}}_{o}^{r} = \mathbf{x}_{o,0}^{r}$, $\hat{\mathbf{x}}_{o}^{h} = \mathbf{x}_{o,0}^{h}$).

For translation, the wrist should be fixed in the root frame ($\hat{\mathbf{x}}_{h}^{r} = \mathbf{x}_{h,0}^{r}$, $\hat{\mathbf{q}}_{h}^{r} = \mathbf{q}_{h,0}^{r}$), and the target object position in the hand frame is dynamically updated by moving the current position along the target direction $\mathbf{d}^{\text{h}}$: $\hat{\mathbf{x}}_{o,t}^{h} = \mathbf{x}_{o,t}^{h} + v_{\max} \Delta t \cdot \mathbf{d}^{\text{h}}$, with $\hat{\mathbf{x}}_{o}^{r}$ derived from the current wrist pose $\mathbf{x}_{h}^{r}$ and $\mathbf{q}_{h}^{r}$. The object orientation maintains initial values in both frames ($\hat{\mathbf{q}}_{o}^{r} = \mathbf{q}_{o,0}^{r}$, $\hat{\mathbf{q}}_{o}^{h} = \mathbf{q}_{o,0}^{h}$).

\subsection{Single-Policy Distillation}
\label{sec:distillation}
Since all per-skill experts share one observation and action space, and each skill is fully specified by its relational targets (\cref{tab:target_poses}), the experts can be straightforwardly distilled into a single cross-skill policy with vanilla DAgger \cite{ross2011dagger}.
 Specifically, we roll out the distilled policy to collect states, query the corresponding per-skill expert actions for the visited states, and train the unified policy on the aggregated state-action pairs using an MSE imitation loss, aiming to learn the desired behavior conditioned on the relational motion objectives.
The skill only selects which expert to query during training and is not provided to the student, which must therefore infer the desired behavior from the relational targets alone.
The environments are uniformly distributed across all valid (object, skill) pairs during training.
More details are provided in SuppMat.

\section{Experiment}
In this section, we first compare our unified policy with per-skill baselines (\cref{sec:per_skill_evaluation}) and evaluate its generalization to unseen objects (\cref{sec:object_generalization}). We further show that our framework applies across different hand morphologies (\cref{sec:cross_morphology}). We then demonstrate the cross-skill capability enabled by our unified framework by long-horizon manipulation (\cref{sec:long_term_manipulation}) and executing relational targets unseen in training (\cref{sec:motion_generalization}). Ablations are finally presented in \cref{sec:ablation}.%

\subsection{Experiment Details}
\label{sec:experiment_setup}

\subsubsection{Objects}
We randomly sample boxes and cylinders in two categories: wrappable objects (all dimensions in $[0.045, 0.075]$\,m), and elongated objects (length in $[0.6, 0.8]$\,m, other dimensions in $[0.045, 0.065]$\,m).
Grasping and relocation are trained and evaluated on both categories, while rotation uses the former and translation the latter.
For each category and object type we sample 200 shapes (100 train / 100 test), yielding 400 training and 400 testing objects.

\subsubsection{Evaluation Setup}
\label{sec:evaluation_setup}

\textbf{Grasping}:
Each object is evaluated with 25 random table-top poses. Fingers initially spread around the object,
with the wrist 5 cm above the object.
With the policy running for 75 time steps, a fixed upward force is applied on the wrist after 50 time steps to lift the object. We report the average success rate (SR (\%)), where success requires the object to be lifted and does not drop throughout the entire episode.

\noindent\textbf{Relocation}:
Each object is evaluated using 25 randomly sampled successful grasping poses. For each trial, a target object pose $\mathbf{T}^\text{final}$ is generated by adding a random positional offset of \([-0.1, 0.1]\) m along the x and y directions, \([0.2, 0.4]\) m along the z direction, and a random angular offset of \([-0.75, 0.75]\) rad for roll, pitch, yaw.
The policy runs for 75 time steps, and we report the average position (PE, cm) and orientation (AE, rad) errors with respect to the target pose over the last 25 time steps of all non-dropping trials. We also report the success rate (SR (\%)), where success requires the object to arrive at the target pose within 3 cm and 0.15 rad without object dropping, respectively.

\noindent\textbf{Rotation \& Translation}:
Each object is evaluated with 25 randomly sampled successful grasping poses for each rotation/translation direction, leading to 150 trials for rotation and 50 trials for translation. The wrist pose is randomly initialized in the world frame.
A trial is considered successful if it achieves an angular displacement exceeding $\pi/2$ rad within 400 time steps or a linear displacement exceeding 5 cm within 50 time steps, without object dropping. We report the success rate (SR (\%)) and the average incremental rotation angle (Rot. (rad)) or translation distance (Trans. (cm)), computed over all non-dropping trials.

\subsection{Per-skill Comparison}
\label{sec:per_skill_evaluation}

We first compare the performance of our unified policy with the per-skill baselines. We choose the RL-in-simulation methods as they are most relevant to our setting. For grasping, we choose GraspXL \cite{zhang2024graspxl}, a method that generates diverse grasping motions for various objects.
For rotation and translation, we choose RotateIt \cite{qi2023general} and \cite{yin2024learninginhandtranslation}, respectively, and adopt their teacher policies with full access to simulated object and hand states for a fair comparison.
For relocation, we utilize the motion synthesis module of D-Grasp \cite{christen2022dgrasp}.
We use Allegro~\cite{Allegro} hand in this section, which is widely used in previous related works.

\begin{table}[t]
    \centering
    \caption{Comparison with Per-skill Baselines under the original settings of the baselines (Ori.) and our general setting (Gen.).}
    \label{tab:comparison}
    \resizebox{1.0\columnwidth}{!}{
        \begin{tabular}{l|c||ccc||cc||cc}
            \toprule
            & Grasp & \multicolumn{3}{c||}{Relocate} & \multicolumn{2}{c||}{Rotate} & \multicolumn{2}{c}{Translate} \\
            \midrule
            Method & SR $\uparrow$ & SR $\uparrow$ & PE $\downarrow$ & AE $\downarrow$ & SR $\uparrow$ & Rot. $\uparrow$ & SR $\uparrow$ & Trans. $\uparrow$ \\
        \midrule
        Ori. (Base.) & 97.5 & -- & -- & -- & 76.5 & 10.4 & 98.3 & 12.8 \\
        Ori. (Ours) & 99.1 & -- & -- & -- & 99.5 & 11.0 & 99.9 & 15.2 \\
        \midrule
        Gen. (Base.) & 97.1 & 92.8 & 0.85 & 0.0231 & 62.7 & 9.43 & 70.6 & 5.81  \\
        Gen. (Ours) & 98.8 & 97.5 & 0.67 & 0.0125 & 98.8 & 10.8 & 99.4 & 15.1 \\
        \bottomrule
        \end{tabular}
    }
    \end{table}

We evaluate under two settings, as reported in \cref{tab:comparison}. We first compare under the original settings of the baselines (Ori.), where grasping is performed without table collision, while rotation and translation are conducted with a fixed wrist-up pose. This aims to demonstrate the advantage of our method under their own favorable conditions. We then adapt the baselines to our more general setting described in \cref{sec:evaluation_setup}, and retrain and re-evaluate them (Gen.), which further reveals their limitations under more general settings. Relocation is only evaluated in the general setting as no prior work addresses it.
Each baseline is trained separately for Ori. and Gen. under the corresponding setting, whereas ours is a single cross-skill policy trained only under our general setting and evaluated in both without retraining.
Additional baseline details are in SuppMat.

Although our single policy is never trained under baseline settings, it consistently outperforms the per-skill baselines that are specifically designed and trained for that setting.
Moving to our more general setting, the rotation and translation baselines drop sharply even after retraining. This is mainly because they rely heavily on finger or palm support rather than stably holding the object. As a result, they degrade sharply once the palm moves away from the upward direction, as the object is much more likely to slip or drop.
In contrast, our policy remains strong in this more difficult general setting, as it starts from grasping poses compatible with subsequent in-hand manipulation and keeps the object stably held throughout the manipulation process.

\subsection{Unseen Object Generalization}
\label{sec:object_generalization}
For a controlled quantitative evaluation of generalization, we construct a test set by randomly sampling 100 spheres and 100 hexagonal prisms with all dimensions in \([0.045, 0.075]\) m, and 200 elongated octagonal prisms with diameter in \([0.045, 0.065]\) m and length in \([0.6, 0.8]\) m. As \cref{tab:generalization} (Unseen (prim.)) shows, despite being trained only on boxes and cylinders, the policy achieves comparable performance on the unseen geometries, indicating strong generalization.

\begin{figure}
	\centering
	\includegraphics[width=1.0\columnwidth]{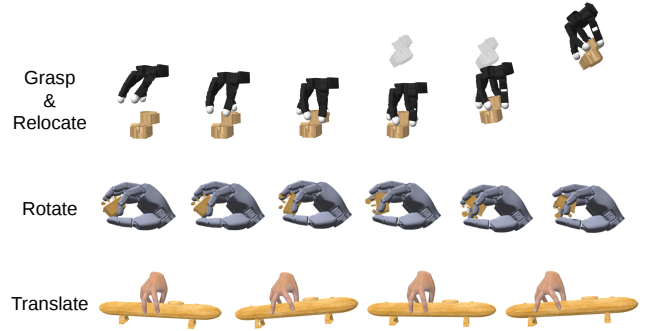}
	\caption{Qualitative results on different hand morphologies and geometrically irregular non-convex unseen objects.}
	\label{fig:unseen_qualitative}
\end{figure}

To further quantify generalization on unseen geometries, we evaluate the policy on the DexYCB objects~\cite{chao2021dexycb}, none of which is used in training.
As many of them are too large for in-hand manipulation (\eg, the pitcher base is about 0.25\,m tall), we uniformly rescale each object to four sizes while preserving their shapes, with the longest side in $[0.07, 0.09]$\,m for rotation and the width in $[0.08, 0.1]$\,m for translation, which are out of the training distribution. The other settings are the same as in \cref{sec:evaluation_setup}, with more details in SuppMat.
As shown in \cref{tab:generalization} (Unseen (YCB)), the performance is relatively lower than on the unseen primitives, mainly due to a few highly challenging objects for in-hand manipulation, \eg, the power drill (6.0\% rotation SR) and the clamp (41.5\% translation SR).
The smaller translation distance arises because these shorter objects do not support large-amplitude translations as they would quickly reach the edges in hand.
Nevertheless, without any additional training, the policy still achieves relatively good performance on these unseen daily objects which are rather different from the training objects.
We provide some qualitative results of other non-convex and geometrically irregular objects in \cref{fig:unseen_qualitative}, with more examples in SuppMat and video.

\begin{table}[t]
    \centering
    \caption{Generalization Evaluation.}
    \label{tab:generalization}
    \resizebox{1.0\columnwidth}{!}{
        \begin{tabular}{l|c||ccc||cc||cc}
            \toprule
            & Grasp & \multicolumn{3}{c||}{Relocate} & \multicolumn{2}{c||}{Rotate} & \multicolumn{2}{c}{Translate} \\
            \midrule
            Setting & SR $\uparrow$ & SR $\uparrow$ & PE $\downarrow$ & AE $\downarrow$ & SR $\uparrow$ & Rot. $\uparrow$ & SR $\uparrow$ & Trans. $\uparrow$ \\
        \midrule
        Allegro & 98.8 & 97.5 & 0.67 & 0.0125 & 98.8 & 10.8 & 99.4 & 15.1 \\
        Unseen (prim.) & 96.5 & 97.1 & 0.57 & 0.0139 & 98.2 & 11.8 & 96.2 & 14.2 \\
        Unseen (YCB) & 81.9 & 81.4 & 1.35 & 0.0280 & 79.7 & 7.68 & 67.9 & 4.72\\
        \midrule
        MANO & 99.2 & 98.2 & 0.96 & 0.0154 & 95.2 & 9.13 & 84.9 & 7.49 \\
        Sharpa & 97.0 & 98.6 & 0.79 & 0.0100 & 98.9 & 11.8 & 99.8 & 12.6 \\
        \bottomrule
        \end{tabular}
}
\end{table}

Notably, leveraging the interaction-aware object representation, the policy exhibits shape- and scale-adaptive manipulation strategies. As shown in the supplementary video, during in-hand rotation, it adapts finger motion patterns to object geometry (e.g., smooth wrapping for round objects vs. edge-aware periodic motions for prisms), and adjusts contact spacing and manipulation gaits across different scales and aspect ratios. These behaviors suggest that the policy learns transferable geometry-adaptive manipulation strategies rather than memorizing object-specific motions.

\subsection{Cross-Morphology Applicability}
\label{sec:cross_morphology}
To verify the applicability of our framework across different hand morphologies, we apply it to another two hands, with expert policies and one unified policy trained for each. The first is MANO \cite{romero2017mano}, a widely used human hand model, for which we build a simulatable hand with 26 degrees of freedom (4 per finger and 6 for the wrist) using the joint definitions from CPF \cite{yang2021cpf}. The second is Sharpa Wave \cite{Sharpa}, an agile dexterous robot hand with five fingers and 28 degrees of freedom (6 for the wrist).
The experimental settings are the same as in \cref{sec:per_skill_evaluation}, except that objects are scaled by 0.7 to match the smaller hand sizes. As shown in \cref{tab:generalization}, while the more limited joint ranges of MANO lead to some performance drops, especially on translation as it requires frequent large finger motions, our framework achieves overall good performance among all three hands, demonstrating its morphology-agnostic applicability.
Qualitative results are shown in \cref{fig:unseen_qualitative}, SuppMat, and video.

\subsection{Long-horizon Manipulation}
\label{sec:long_term_manipulation}
With cross-skill capability and state compatibility captured by the unified formulation, our unified policy naturally supports long-horizon manipulation by chaining different skills, which is infeasible for prior per-skill methods due to their skill-specific settings.
We evaluate this capability on the two groups of test objects with sequences of either grasp + relocate + rotate (Gr. + Re. + Rot.) or grasp + relocate + translate (Gr. + Re. + Trans.).
Stage transitions update the relational targets while keeping the same policy active.
Each test object is evaluated with 25 random poses on the table. The policy continuously executes a seamless sequence: starting from the initial pose, it grasps the object, relocates it to a target pose randomly sampled within the same ranges as in \cref{sec:evaluation_setup}, and continues to perform in-hand rotation or translation along a randomly selected direction, all in a single continuous execution.
A sequence is successful only if all three phases succeed, including grasping (the object is lifted without dropping), relocation (the object arrives at the target pose within 3~cm and 0.15~rad), and in-hand manipulation (rotation or translation exceeds $\pi/2$~rad or 5~cm respectively without object dropping). Per-phase success rates are conditioned on the success of all preceding phases.

\begin{table}[t]
\centering
\caption{Long-horizon Manipulation.}
\label{tab:long_term}
\resizebox{1.0\columnwidth}{!}{

\begin{tabular}{l|ccc|c}
    \toprule
    Sequence & Grasp SR $\uparrow$ & Relocate SR $\uparrow$ & In-hand SR $\uparrow$ & Overall SR $\uparrow$ \\
    \midrule
    Gr. + Re. + Rot. & 99.6 & 95.4 & 95.3 & 90.5 \\
    Gr. + Re. + Trans. & 99.9 & 98.5 & 95.9 & 94.4 \\
    \bottomrule
\end{tabular}
}
\end{table}

We report the per-phase and overall success rates in \cref{tab:long_term}.
The in-hand (third phase) success rates are relatively lower than in the single-skill evaluation in \cref{tab:comparison} because, after relocation, the hand starts from downward or diagonally downward orientations rather than from randomly sampled ones, which is inherently more challenging as the object is more prone to slipping or dropping.
In addition, since relocation starts immediately after grasping, residual motion from the grasping phase can perturb the subsequent relocation, which explains the slight relocation performance drop observed on the smaller wrappable objects used in the rotation sequences.
Nevertheless, our single policy achieves good overall performance.
Some qualitative results are shown in \cref{fig:teaser}, with more demonstrations in SuppMat and video.

\subsection{Unseen Relational Targets}
\label{sec:motion_generalization}
Beyond temporal skill chaining, our unified formulation which instantiates different skills with continuous relational targets rather than discrete skill identifiers also helps with control compatibility, where the policy can deal with the relational targets that are unseen during training. To evaluate this capability, we test its in-hand rotation performance about randomly sampled axes on the unit sphere, with all other settings unchanged (\cref{sec:evaluation_setup}).
The policy achieves a 93.8\% SR (\vs 98.8\% on the canonical axes in \cref{tab:comparison}), remaining above 98\% within 30\textdegree{} of the nearest canonical axis and 87.1\% at the largest possible deviations (40\textdegree{}--54.7\textdegree{}).
This indicates that behaviors learned for different targets are modeled in a shared solution space and are compatible enough to be smoothly bridged.

\begin{table}[t]
    \centering
    \caption{Ablation Study.}
    \label{tab:ablation}
    \resizebox{1.0\columnwidth}{!}{%
    \begin{tabular}{l|c||ccc||cc||cc}
        \toprule
         & Grasp & \multicolumn{3}{c||}{Relocate} & \multicolumn{2}{c||}{Rotate} & \multicolumn{2}{c}{Translate} \\
        \midrule
        Method & SR $\uparrow$ & SR $\uparrow$ & PE $\downarrow$ & AE $\downarrow$ & SR $\uparrow$ & Rot. $\uparrow$ & SR $\uparrow$ & Trans. $\uparrow$ \\
        \midrule
        W/o Obs. & 98.5 & 97.6 & 0.65 & 0.0139 & 0.00 & 0.07 & 4.81 & 1.84 \\
        W/o Rew. & 98.7 & 61.3 & 5.24 & 0.0180 & 97.4 & 13.2 & 99.1 & 13.7 \\
        W/o Act. & 97.8 & 37.4 & 9.03 & 0.1723 & 96.5 & 11.7 & 99.3 & 14.4 \\
        \midrule
        Per-skill & 99.0 & 98.8 & 0.59 & 0.0138 & 99.1 & 12.3 & 99.6 & 15.3 \\
        \midrule
        Ours & 98.8 & 97.5 & 0.67 & 0.0125 & 98.8 & 10.8 & 99.4 & 15.1 \\
        \bottomrule
        \end{tabular}%
    }
\end{table}

\subsection{Ablation}
\label{sec:ablation}

Our relation-driven formulation shares unified observation, action, and reward structures, in which the components conditioned on the target relational motion objectives play an important role.
To verify their contribution, we ablate the target pose observation $\mathbf{g}_t^{\text{target}}$ (W/o Obs.), the tracking reward $r_t^{\text{track}}$ (W/o Rew.), and the residual wrist action (W/o Act.), where the wrist $\mathbf{q}_{t}^{\text{ref}}$ is set to $\mathbf{q}_{t-1}^{\text{act}}$ as for the finger joints. Each variant retrains its experts before distillation.

As shown in \cref{tab:ablation}, the target-conditioned components affect different skills in distinct ways.
Without the target pose observation $\mathbf{g}_t^{\text{target}}$, the policy completely fails on rotation and translation.
This demonstrates the necessity of the unified target pose observation.
More importantly, it shows that the unified cross-skill policy effectively infers the desired behavior from the target relational motion objectives including the target pose observation.
W/o Act.\ causes the largest relocation degradation as it impairs efficient wrist exploration for large-amplitude motions.
W/o Rew.\ also harms relocation, which relies on dense pose-tracking supervision, while other skills not requiring exact pose matching degrade less since $r_t^{\text{goal}}$ still provides interaction objectives. The higher rotation angle (Rot.) of W/o Rew.\ results from more aggressive rotation without $r_t^{\text{track}}$ regulating the speed around $\omega_{\max}$, at the cost of a lower SR.
These ablations validate the effectiveness and necessity of our unified formulation defined from the relational motion perspective.

To further demonstrate the effectiveness of the formulation, we compare our unified policy with the per-skill policies it is distilled from (Per-skill in \cref{tab:ablation}). Our policy closely matches the experts in success rate on all skills, conditioned only on current states and target relational poses without explicit skill identifiers. For long-horizon manipulation, switching among the per-skill experts, which requires explicit policy selection for every stage, also yields comparable overall SRs (92.3 and 94.7 for the rotation and translation sequences, \vs 90.5 and 94.4 of ours in \cref{tab:long_term}).
This close match indicates the compatibility of behaviors learned under the shared formulation, while the straightforward single policy distillation serves as a natural consequence of the conpatibility, and further reduces the need for policy switching during inference.

\section{Conclusion and Future Work}
In this work, we presented a unified framework for cross-skill dexterous manipulation that models grasping, relocation, in-hand rotation, and in-hand translation from the hand-object relational perspective, leading to a unified formulation with shared state spaces, action spaces, and reward structures.
The formulation supports cross-skill compatibility, allowing straightforward distillation of a single cross-skill policy capable of long-horizon manipulation based on relational targets.
Experiments show the cross-skill policy achieves strong per-skill performance and generalization to unseen geometries with adaptive behaviors.
The framework also supports different hand morphologies.
Moreover, the policy can execute relational targets beyond the training distribution, suggesting a shared solution space captured by our framework.

This work focuses on four canonical skills as a first step to demonstrate and exploit the cross-skill compatibility and the shared solution space.
Many everyday dexterous motions, however, cannot be clearly attributed to any specific skill. As such motions may also be expressed as relational targets, exploring them within the shared solution space beyond explicit skill boundaries is an interesting next step toward general-purpose hand-object motion synthesis.
Nevertheless, our results provide preliminary evidence that seemingly distinct dexterous manipulation skills, viewed from an appropriate perspective, has the potential to be unified within a shared solution space as different instantiations.

{
    \small
    \bibliographystyle{ieeenat_fullname}
    \bibliography{sample-bibliography}
}

\clearpage
\makeatletter
\hypersetup{plainpages=true}
\edef\supp@pageoffset{\the\numexpr\c@page-\@ne\relax}
\renewcommand{\thepage}{\number\numexpr\c@page-\supp@pageoffset\relax}
\setcounter{figure}{0}
\setcounter{table}{0}
\setcounter{equation}{0}
\setcounter{footnote}{0}
\renewcommand{\theHfigure}{supp.\arabic{figure}}
\renewcommand{\theHtable}{supp.\arabic{table}}
\renewcommand{\theHequation}{supp.\arabic{equation}}
\gdef\@extra@b@citeb{@supp}
\let\supp@NAT@wrout\NAT@wrout
\def\NAT@wrout#1#2#3#4#5{\supp@NAT@wrout{#1}{#2}{#3}{#4}{#5\@extra@b@citeb}}
\renewcommand*{\backrefxxx}[3]{\hyperlink{page.\the\numexpr#1+\supp@pageoffset\relax}{#1}}
\makeatother

\maketitlesupplementary
\appendix
The supplementary material includes this PDF and a video, which show qualitative results of our method.

\section{Implementation Details}
As explained in Sec. 3, we use PPO \cite{schulman2017proximal} to train the per-skill policies, and DAgger \cite{ross2011dagger} to distill them into a single cross-skill policy. The policies are composed of one MLP encoder for the objective features, and one main MLP that predicts the actions with $\mathbf{s}_t^{\text{h}}, \mathbf{s}_t^{\text{o}}$ and the output of the encoder as input. The encoder has 2 hidden layers with dimensions [128, 64], while the main MLP has 3 hidden layers with dimensions [512, 256, 128].
We present the PPO hyperparameters in \cref{tab:params}, and the DAgger hyperparameters in \cref{tab:dagger_params}. We further provide the reward coefficients in \cref{tab:reward_coeffs}.
All policies are trained on a single NVIDIA GeForce RTX 4090. Each per-skill expert takes 6 hours on average, and distilling the experts into the unified policy takes 12 hours.

\begin{table}[htb]
    \caption{Hyperparameters of PPO.}
    \label{tab:params}
    \centering
    \resizebox{0.65\columnwidth}{!}{
    \begin{tabular}{ll}
    \toprule
    \textbf{Hyperparameter} & \textbf{Value} \\
    \midrule
    Discount factor $\gamma$ & 0.99 \\
    GAE parameter $\lambda$ & 0.95 \\
    Learning rate & 0.005 \\
    KL threshold & 0.02 \\
    Truncate gradients & True \\
    Max. gradient norm & 1.0 \\
    Mini epochs & 5 \\
    Batch size & 32768 \\
    \bottomrule
    \end{tabular}
    }
\end{table}

\begin{table}[htb]
    \caption{Hyperparameters of DAgger.}
    \label{tab:dagger_params}
    \centering
    \resizebox{0.75\columnwidth}{!}{
    \begin{tabular}{ll}
    \toprule
    \textbf{Hyperparameter} & \textbf{Value} \\
    \midrule
    Batch size & 16384 \\
    Learning rate & 3.0e-4 \\
    Max. epochs & 3500 \\
    Gradient updates per epoch & 100 \\
    Rollout steps & 32 \\
    Replay buffer size & 1.0e6 \\
    Expert query rate & 100\% \\
    \bottomrule
    \end{tabular}
    }
\end{table}

\begin{table}[htb]
    \caption{Reward coefficients.}
    \label{tab:reward_coeffs}
    \centering
    \resizebox{1.0\columnwidth}{!}{
    \begin{tabular}{lll}
    \toprule
    \textbf{Reward Term} & \textbf{Reward Coefficient} & \textbf{Value} \\
    \midrule
    \multirow{5}{*}{\shortstack{\textit{Interaction} \\  \textit{Objective} \\  \textit{($r_t^{\text{goal}}$)}}} & $w_{\text{dis}}$ & 5.0\\
    & $w_{\text{con}}$ & 0.2\\
    & $w_{\text{f}}$ & 0.02\\
     & $w_{\text{p, linear}}$  & 10.0\\
     & $w_{\text{p, angular}}$  & 1.0\\
    \midrule
    \multirow{6}{*}{\shortstack{\textit{Target Tracking} \\ \textit{($r_t^{\text{track}}$)}}} & $w_{opr}$ (object position, root frame) & -5.0\\
     & $w_{oqr}$ (object quaternion, root frame) & -1.0\\
     & $w_{oph}$ (object position, hand frame) & -5.0\\
     & $w_{oqh}$ (object quaternion, hand frame) & -1.0\\
     & $w_{hpr}$ (hand position, root frame) & -15.0\\
     & $w_{hqr}$ (hand quaternion, root frame) & -3.0\\
    \midrule
    \multirow{5}{*}{\shortstack{\textit{Regularization} \\ \textit{($r_t^{\text{reg}}$)}}} & $w_{\text{pose}}$ (finger pose penalty) & -0.3\\
     & $w_{\text{vel}}$ (wrist translational velocity) & -0.5\\
     & $w_{\text{ang}}$ (wrist angular velocity) & -0.05\\
     & $w_{\tau}$ (torque magnitude) & -0.1\\
     & $w_{\text{drop}}$ (drop penalty) & -10.0\\
    \bottomrule
    \end{tabular}
    }
\end{table}

\section{Baseline Details}
Even though there are some previous works that address dexterous manipulation with isolated skills, they usually have different focuses or run under more limited settings compared with ours. As a result, we adapt the baselines for the general setting evaluation.

GraspXL \cite{zhang2024graspxl} focuses on generating grasping motions with diverse approach directions regardless of table collision. To adapt it for our setting, we apply the top-down approach direction and add a table collision penalty reward term during training for a fair comparison.

\begin{figure*}
	\centering
	\includegraphics[width=0.93\textwidth]{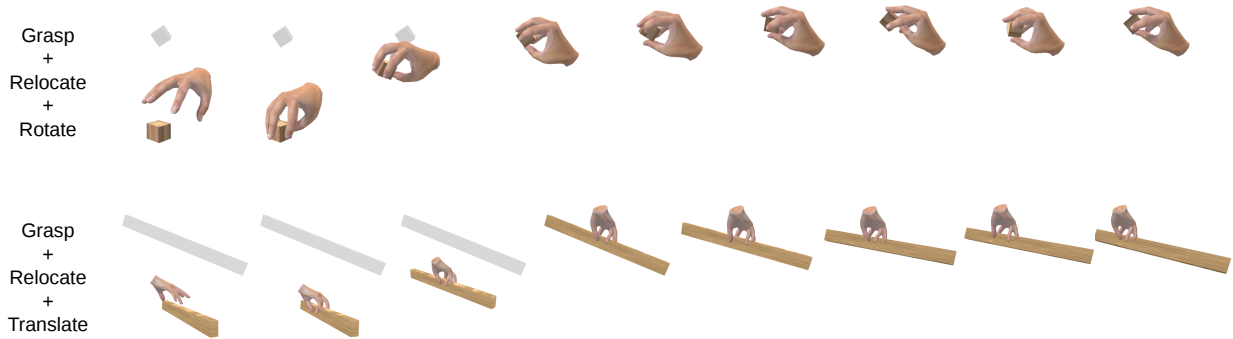}
	\caption{Human Hand long-horizon manipulation}
	\label{fig:mano_long_term}
	\vspace{7mm}
\end{figure*}

\begin{figure*}
	\centering
	\includegraphics[width=0.92\textwidth]{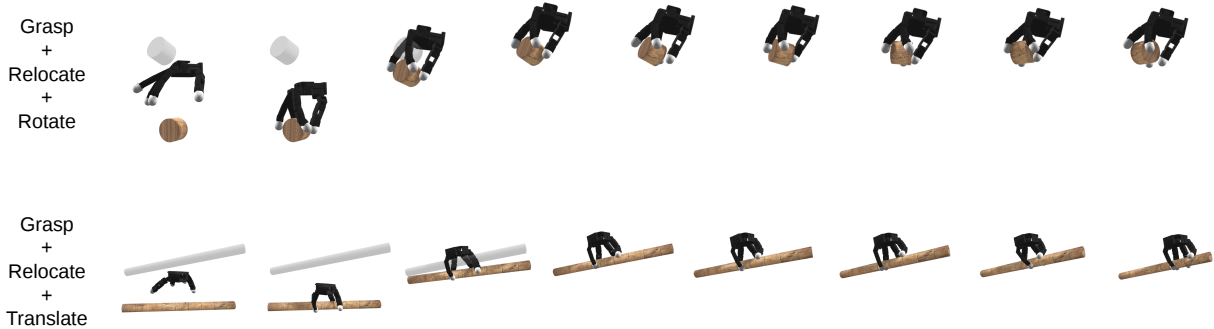}
	\caption{Allegro long-horizon manipulation}
	\label{fig:allegro_long_term}
	\vspace{7mm}
\end{figure*}

\begin{figure*}[!ht]
	\centering
	\includegraphics[width=0.93\textwidth]{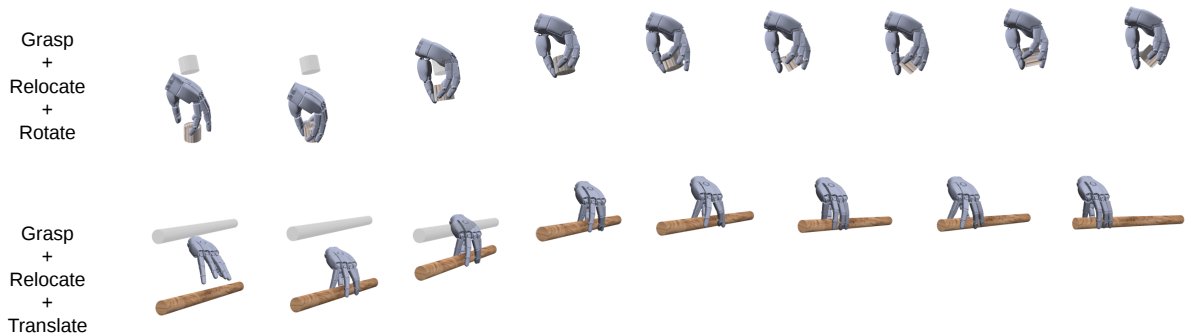}
	\caption{Sharpa Wave long-horizon manipulation}
	\label{fig:sharpa_long_term}
    \vspace{7mm}
\end{figure*}

The rotation baseline RotateIt \cite{qi2023general} and the translation baseline \cite{yin2024learninginhandtranslation} are both designed for a hand-up setting with fixed wrists, and they highly rely on the palm support to stabilize the object.
For a fair comparison, we utilize their teacher policy as the baseline. For \cite{yin2024learninginhandtranslation}, specifically, as it encourages translation to a specific target position instead of continuous translation, we change the reward to encourage translation along the target axis.

\section{Additional Qualitative Results}
We provide qualitative long-horizon manipulation results for the MANO, Allegro, and Sharpa Wave hands in \cref{fig:mano_long_term,fig:allegro_long_term,fig:sharpa_long_term}. More qualitative results are provided in the video.

\section{DexYCB Evaluation}
\label{sec:supp_dexycb}
Objects whose longest side is at least twice their shortest side are used for translation (5 objects), and the remaining objects are used for rotation. Grasping and relocation use all objects.
Each object is uniformly scaled to four sizes with their shapes preserved.
For rotation, the longest side takes four evenly spaced values in $[0.07, 0.09]$\,m, with the other sides accordingly scaled to $[0.022, 0.088]$\,m.
For translation, the second-longest side takes four evenly spaced values in $[0.08, 0.1]$\,m, resulting in lengths in $[0.11, 0.52]$\,m.
In comparison, the training wrappable objects have all dimensions in $[0.045, 0.075]$\,m, and the training elongated objects have lengths in $[0.6, 0.8]$\,m and other dimensions in $[0.045, 0.065]$\,m.
Each scaled object is evaluated as one test object following the same procedure as in other experiments, with grasping evaluated under 25 random table-top poses, and relocation, rotation, and translation starting from 25 randomly sampled successful grasping poses collected on the same scaled object.

\end{document}